%% file: main.tex
\newcommand{\CLASSINPUTtoptextmargin}{0.75in}
\newcommand{\CLASSINPUTbottomtextmargin}{0.75in}
\newcommand{\CLASSINPUTinnersidemargin}{0.75in}
\newcommand{\CLASSINPUToutersidemargin}{0.75in}
\documentclass[letterpaper, 10 pt, conference]{ieeetran}  

\IEEEoverridecommandlockouts                              

\usepackage[utf8]{inputenc}
\usepackage[T1]{fontenc}    
\usepackage{multicol}
\usepackage{hyperref}
\hypersetup{bookmarksopen,bookmarksnumbered,
pdfpagemode=UseOutlines,
colorlinks=true,
linkcolor=blue,
anchorcolor=blue,
citecolor=blue,
filecolor=blue,
menucolor=blue,
urlcolor=blue
}
\usepackage{graphicx}
\usepackage{amsmath}
\usepackage{upgreek}
\usepackage{amsfonts,amssymb}
\usepackage{bm}
\usepackage{bbm}

\usepackage{amsthm} 
\usepackage{thmtools}
\usepackage{mathtools}
\usepackage{epstopdf}
\usepackage{xspace}
\usepackage[numbers]{natbib}

\usepackage[font=footnotesize]{caption} 
\usepackage[font=footnotesize]{subcaption}
\usepackage{units}
\usepackage{booktabs} 
\usepackage{tabulary}
\usepackage[usenames,dvipsnames]{color} 
\usepackage{diagbox}
\usepackage[ruled,vlined,linesnumbered]{algorithm2e}  
\usepackage{parskip}
\SetKwComment{Comment}{$\triangleright$\ }{}
\usepackage{ifthen,version}
\usepackage{soul}
\usepackage{fixltx2e}
\usepackage{csquotes}
\usepackage{microtype}      
\usepackage{lipsum}
\usepackage{csquotes}
\usepackage{tikz}
\let\labelindent\relax
\usepackage{enumitem}

\newcommand{\fullFigGap}[0]{\vspace{-1.5\baselineskip}} 
\newcommand{\red}[0]{\color{red}}

\newcommand{\xxnote}[3]{}
\ifx\hidenotes\undefined
  \usepackage{color}
  \renewcommand{\xxnote}[3]{\color{#2}{#1: #3}}
\fi

\newtheoremstyle{hypstyle}
{3pt} 
{3pt} 
{\itshape} 
{} 
{\bfseries} 
{.} 
{.5em} 
{} 

\theoremstyle{hypstyle} 
\newtheorem{observation}{O}
\newtheorem{ques}{Q}

\newcommand{\algorithmStyle}[0]{\footnotesize}

\input{math_operators}

\input{math_defns}

\input{text_shortcuts}

\title{\LARGE \bf
Bayesian Active Edge Evaluation on Expensive Graphs
}

\author{Sanjiban Choudhury$^{1}$, Siddhartha Srinivasa$^{2}$ and Sebastian Scherer$^{1}$
\thanks{$^{1}$S.\ Choudhury and S.\ Scherer are with The Robotics Institute, Carnegie Mellon University, USA {\tt\small \{sanjiban, basti\}@cmu.edu}}%
\thanks{$^{2}$S.\ Srinivasa is with School of Computer Science and Engineering, University of Washington, USA {\tt\small \{siddh\}@cs.uw.edu}}
}
\begin{document}

\maketitle
\thispagestyle{empty}
\pagestyle{empty}

\input{abstract}

\input{introduction}

\input{problem_formulation}

\input{related_work}

\input{approach}

\input{experiments}

\input{conclusion}

\section{Acknowledgement}
The authors thank Shushman Choudhury for feedback, insightful discussions and the robot arm dataset. They also thank Shervin Javdani for helpful tips on \direct implementation.

\footnotesize{
\bibliographystyle{plainnat}
\bibliography{reference}
}
\end{document}

%% file: math_operators.tex
\DeclareMathOperator*{\argmin}{arg\,min}
\DeclareMathOperator*{\argmax}{arg\,max}

\newcommand{\argmaxprob}[1]{\underset{#1}{\argmax}}
\newcommand{\argminprob}[1]{\underset{#1}{\argmin}}

\newcommand{\abs}[1]{\left|#1 \right|}

\newcommand{\expect}[2]{\mathbb{E}_{#1}\left[#2\right]}

\newcommand{\real}[0]{\mathbb{R}}

\newcommand{\bbm}{\begin{bmatrix}}
\newcommand{\ebm}{\end{bmatrix}}

\newcommand{\pair}[2]{\left( #1, #2\right)}
\newcommand{\set}[1]{\left\lbrace #1\right\rbrace}
\newcommand{\seq}[2]{\left(#1_{1}, #1_{2}, \ldots, #1_{#2}\right)}

\newcommand{\setst}[2]{\left\lbrace #1\;\middle|\;#2\right\rbrace}

%% file: math_defns.tex
\newcommand{\explicitGraph}[0]{G}
\newcommand{\vertexSet}[0]{V}
\newcommand{\edgeSet}[0]{E}
\newcommand{\edge}[0]{e}
\newcommand{\start}[0]{v_s}
\newcommand{\goal}[0]{v_g}
\newcommand{\Path}[0]{\xi}
\newcommand{\PathSet}[0]{\Xi}
\newcommand{\world}[0]{\phi}

\newcommand{\selTestSet}[0]{\mathcal{A}}

\newcommand{\outcome}[0]{x}
\newcommand{\outcomeSet}[1]{\mathbf{\outcome}_{#1}}
\newcommand{\obsOutcome}[0]{\outcomeSet{\selTestSet}}
\newcommand{\obsOutcomeFunc}[2]{\obsOutcome \left( #1, #2\right)}

\newcommand{\test}[0]{t}
\newcommand{\testSet}[0]{\mathcal{T}}

\newcommand{\outcomeTest}[1]{\outcome_{#1}}
\newcommand{\obsOutcomeAdd}[1]{\outcomeSet{\selTestSet \cup \set{#1}}}

\newcommand{\bias}[0]{\theta}
\newcommand{\biasVec}[0]{\bm{\bias}}

\newcommand{\region}[0]{\mathcal{R}}
\newcommand{\numRegion}[0]{m}

\newcommand{\policy}[0]{\pi}

\newcommand{\cost}[0]{c}
\newcommand{\policyOpt}[0]{\pi^*}

\newcommand{\graphEC}[0]{\mathcal{G}_\mathrm{EC}}
\newcommand{\vertexEC}[0]{\mathcal{V}_\mathrm{EC}}

\newcommand{\hyp}[0]{h}
\newcommand{\hypSet}[0]{\mathcal{H}}
\newcommand{\hypSpace}[0]{\mathcal{H}}

\newcommand{\numHyp}[0]{n}

\newcommand{\versionSpace}[0]{\hypSet}

\newcommand{\edgeSetEC}[0]{\mathcal{E}_\mathrm{EC}}
\newcommand{\edgeFnEC}[2]{\{#1, #2\}}

\newcommand{\weight}[0]{w}

\newcommand{\ec}[0]{f_\mathrm{EC}}
\newcommand{\fec}[1]{f_\mathrm{EC}(#1)}

\newcommand{\gain}[3]{\Delta_{#1} \left( #2 \;|\; #3 \right)}

\newcommand{\wec}[0]{w_\mathrm{EC}}

\newcommand{\drd}[0]{f_\mathrm{DRD}}
\newcommand{\fdrd}[1]{f_\mathrm{DRD}(#1)}
\newcommand{\feci}[2]{f_\mathrm{EC}^{#1}(#2)}

\newcommand{\bigo}[1]{\mathcal{O}\left(#1\right)}

\newcommand{\gainTest}[0]{\Delta}
\newcommand{\activeHypSet}[0]{\hypSet_\mathrm{act}}
\newcommand{\hypRegion}[0]{\mathbf{R}}
\newcommand{\hypTest}[0]{\mathbf{X}}
\newcommand{\costTest}[0]{\mathbf{c}}
\newcommand{\condActiveHypSet}[0]{\hypSet_\mathrm{cond}}
\newcommand{\probTest}[0]{p}
\newcommand{\hypSetTwo}[0]{\hypSet'}

\newcommand{\threshDIRECT}[0]{\eta}
\newcommand{\versionSpaceDIRECT}[0]{\hypSet_\threshDIRECT}
\newcommand{\mixDIRECT}[0]{\alpha}

%% file: text_shortcuts.tex
\newcommand{\etal}[0]{et al.}

\newcommand{\bisect}[0]{\textsc{BiSECt}\xspace}
\newcommand{\direct}[0]{\textsc{DiRECt}\xspace}
\newcommand{\directONLY}[0]{\textsc{DiRECtonly}\xspace}

\newcommand{\ecsq}[0]{EC\textsuperscript{2}\xspace}

\newcommand{\algMaxTally}[0]{\textsc{MaxTally}\xspace}
\newcommand{\algSetCover}[0]{\textsc{SetCover}\xspace}
\newcommand{\algMVOI}[0]{\textsc{MVoI}\xspace}

\newcommand{\algMaxProbReg}[0]{\textsc{MaxProbReg}\xspace}

\newcommand{\algLazySP}[0]{\textsc{LazySP}\xspace}
\newcommand{\algLazySPSet}[0]{\textsc{LazySPSet}\xspace}

%% file: abstract.tex
\begin{abstract}
Robots operate in environments with varying implicit structure. 
For instance, a helicopter flying over terrain encounters a very different arrangement of obstacles than a robotic arm manipulating objects on a cluttered table top. 
State-of-the-art motion planning systems do not exploit this structure, thereby expending valuable planning effort searching for implausible solutions. 
We are interested in planning algorithms that \emph{actively infer the underlying structure} of the valid configuration space during planning in order to find solutions with minimal effort. \\
Consider the problem of evaluating edges on a graph to quickly discover collision-free paths. 
Evaluating edges is expensive, both for robots with complex geometries like robot arms, and for robots with limited onboard computation like UAVs.
Until now, this challenge has been addressed via \emph{laziness} i.e. deferring edge evaluation until absolutely necessary, with the hope that edges turn out to be valid. 
However, all edges are not alike in value - some have a lot of potentially good paths flowing through them, and some others encode the likelihood of neighbouring edges being valid.
This leads to our key insight - instead of passive laziness, we can \emph{actively} choose edges that reduce the uncertainty about the validity of paths. 
We show that this is equivalent to the Bayesian active learning paradigm of \emph{decision region determination (DRD)}.
However, the DRD problem is not only combinatorially hard, but also requires explicit enumeration of all possible worlds. 
We propose a novel framework that combines two DRD algorithms, \direct and \bisect, to overcome both issues.
We show that our approach outperforms several state-of-the-art algorithms on a spectrum of planning problems for mobile robots, manipulators and autonomous helicopters.

\end{abstract}

%% file: introduction.tex
\section{Introduction}
\label{sec:intro}

\begin{figure}[!t]
    \centering
    \includegraphics[page=1,width=\columnwidth]{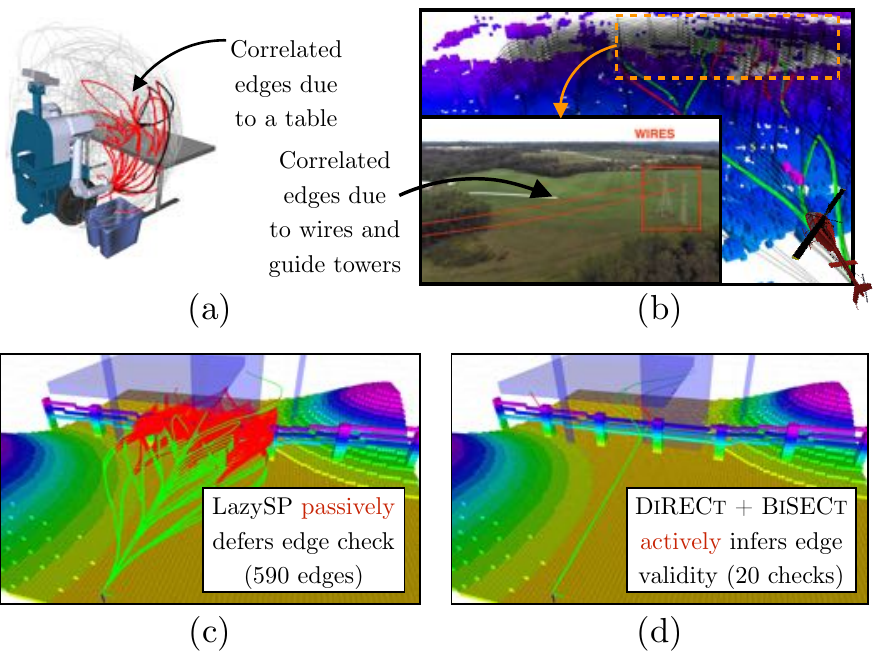}
    \caption{%
    \label{fig:real_world_planning}
    Real world planning problems where edges are correlated. In such cases, our approach can infer the structure of the world from outcomes of edge evaluations.  (a) The presence of a table in robotic arm planning correlates neighbouring edges (courtesy Dellin~\cite{dellin2016unifying}). (b) The presence of wires and guide-towers in helicopter planning correlates corresponding edges.
    (c) A typical helicopter planning problem with wires, terrain and no-fly zones. The state-of-the-art planner, LazySP \cite{dellin2016unifying}, passively defers edge evaluation thus requiring 590 checks. It is unable to leverage priors on the world.
    (d) Our approach uses \direct to actively infer the presence of the wire, hills and NFZ and \bisect to focus the search and find a path in 20 checks.
    }
\end{figure}

A widely-used approach for solving robot motion-planning problems is the construction of graphs, where vertices represent robot configurations and edges represent potentially valid movements of the robot between these configurations.
The main computational bottleneck is collision checking which is manifested as expensive edge evaluations.
For example, in robot arm planning~\cite{dellin2016guided} (Fig.~\ref{fig:real_world_planning}(a)), evaluation requires expensive geometric intersection computations. In autonomous helicopter planning~\cite{choudhury2014planner} (Fig.~\ref{fig:real_world_planning}(b)), evaluation requires expensive reachability volume verification of the closed loop system. 
State-of-the-art planning algorithms~\cite{dellin2016unifying} deal with expensive evaluation by resorting to \emph{laziness} - they first compute a set of unevaluated paths quickly, and then evaluate them sequentially until a valid path is found. 


\begin{figure*}[!t]
    \centering
    \includegraphics[width=\textwidth]{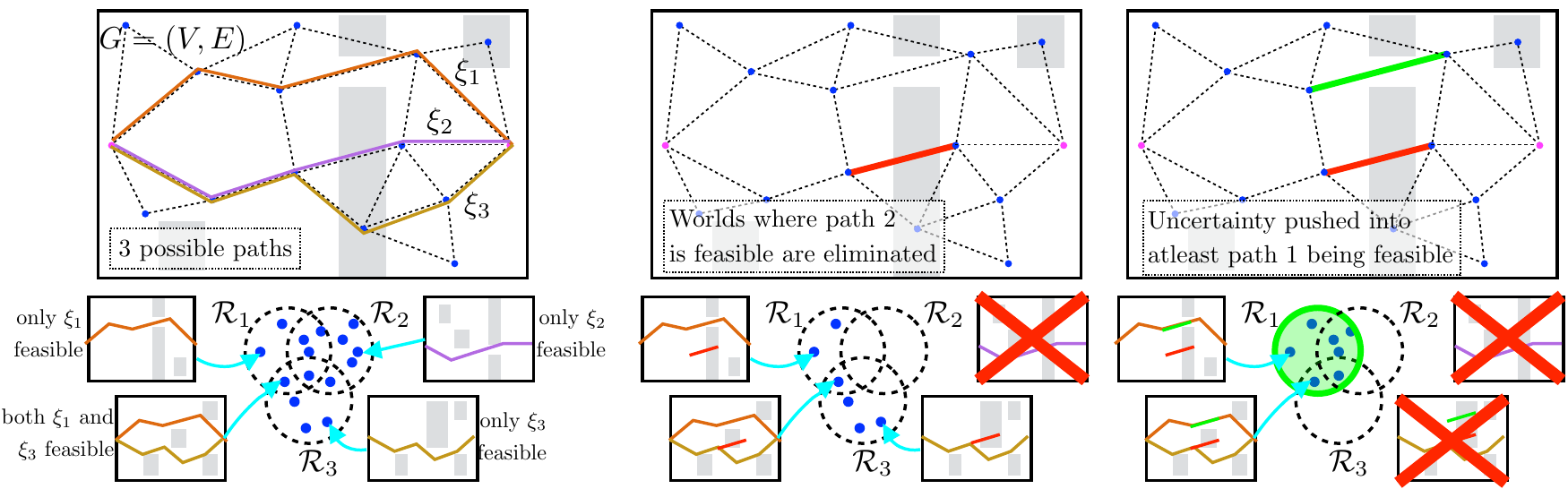}
    \caption{%
    \label{fig:general_drd_problem}
    Equivalence between the feasible path identification problem and the decision region determination problem. 
    A plausible world is equivalent to hypothesis (as shown by the blue dots in the lower row).
    A path $\Path_i$ is equivalent to a region $\region_i$ over valid hypotheses where the path is feasible.
    A collision check is equivalent to a test whose outcome is valid (green) or invalid (red). 
    Tests eliminate hypotheses and the algorithm terminates when uncertainty is pushed into a region ($\region_1$) and the corresponding path ($\Path_1$) is determined to be valid. \fullFigGap
    }
\end{figure*}%

However, such lazy policies overlook a fundamental characteristic of the planning problem - \emph{edges in a graph are implicitly correlated}. 
In Fig.~\ref{fig:real_world_planning}(a), the presence of a table in the robot arm workspace implicitly correlates edges in front of the robot. 
Similarly in Fig.~\ref{fig:real_world_planning}(b), the presence of power-lines during a UAV overflight implicitly correlates a horizontal strip of edges near the ground. 
Evaluating such edges provides valuable information about the feasibility likelihood of other edges which in turn can be used to infer the feasibility likelihood of a path.
We wish to compute such a policy that judiciously chooses edges to evaluate by reasoning about likely worlds in which the robot operates.

This problem is equivalent to the Bayesian active learning problem of \emph{decision region determination (DRD)}~\cite{javdani2014near, chen2015submodular} - given a set of tests (edges), hypotheses (worlds), and regions (potential paths), the objective is to select a sequence of tests that drive uncertainty into a single decision region. The DRD problem has one key distinction from the general active learning problem \cite{dasgupta2004analysis} - we only need to know enough about the world to ascertain if a path is feasible.  

To solve the DRD problem in our context, we need to address two issues:
\begin{enumerate}[label=(\alph*)]
\item\label{pt1} Enumeration of all possible worlds.
\item\label{pt2} Solving the DRD problem in general is NP-hard \cite{javdani2014near}.
\end{enumerate}
Fortunately, Chen \etal~\cite{chen2015submodular} provide an algorithm, \direct, to address \ref{pt2} by maximizing an objective function that satisfies \emph{adaptive submodularity}~\cite{golovin2011adaptive} - a natural diminishing returns property that endows greedy policies with near-optimality guarantees. 

However, \direct requires \ref{pt1} to be solved, i.e. requires an exhaustive training database of worlds. Since \direct operates on a realizability assumption, it can easily terminate without finding a solution when the test world is not in its training database. Explicitly enumerating all possible worlds is impractical even as an offline operation - a graph with $E$ edges can induce $\bigo{2^E}$ possible worlds.\footnote{A typical graph, $\abs{E}: 10000$, will need $2^{10000}$ bits of storage!} 

In previous work~\cite{choudhury2017active}, we addressed \ref{pt1} by examining the DRD problem when edges are \emph{independent}. We proposed an efficient near-optimal algorithm \bisect which reduces the computation from $\bigo{2^E}$ to $\bigo{E}$. \bisect reasons about the exhaustive set of worlds without ever explicitly enumerating them by leveraging the independence assumption. However, this assumption is too strong for certain environments (such as those in Fig~\ref{fig:real_world_planning}) thus leading to excessive edge evaluations. 

Our key idea is to combine the two approaches. We sample a finite database of worlds and apply \direct offline on this database to compute a decision tree of edges to evaluate. At test time we execute the tree. When we reach a leaf node, we have either solved the problem or we have narrowed the problem down to a set of `tail worlds' outside of \direct's domain, i.e. low probability worlds that do not appear in the sampled database. We then run \bisect, which implicitly reasons about this set of `tail worlds', and accept the performance loss due to the independence assumption. We make the following contributions:
\begin{enumerate}
  \item We show an equivalence between the optimal edge evaluation problem and the decision region determination problem.
  \item We propose a framework to combine two DRD algorithms, \direct and \bisect, that near-optimally solves the decision region problem, overcomes issues pertaining to finite databases and can be executed efficiently online.
  \item We demonstrate the efficacy of our approach on a spectrum of planning problems for mobile robots, manipulators, autonomous full-scale helicopters. 
\end{enumerate}
We note that a limitation of this approach is that it ignores solution quality and requires an explicit library of paths. We discuss ways to alleviate this in Section~\ref{sec:conclusion}.

%% file: problem_formulation.tex
\section{Problem Formulation}
\label{sec:problem_formulation}

We now describe the edge evaluation problem, showing the equivalence to the DRD problem along the way.
Let $\explicitGraph = \pair{\vertexSet}{\edgeSet}$ be an explicit graph that consists of a set of vertices $\vertexSet$ and edges $\edgeSet$. Given a pair of start and goal vertices, $\pair{\start}{\goal} \in \vertexSet$, a search algorithm computes a path $\Path \subseteq \edgeSet$ - a connected sequence of valid edges. 
The search is performed on an underlying world $\world$ which corresponds to a specific setting of obstacles.
To ascertain the validity of an edge $\edge$, the algorithm queries the underlying world $\world(\edge)$ which returns a binary status. We address applications where edge evaluation is expensive, i.e., the computational cost $\cost(\edge)$ of computing $\world(\edge)$ is significantly higher than regular search operations. We make a simplification to the problem - from that of search to that of identification. Instead of searching $\explicitGraph$ online for a path, we frame the problem as identifying a valid path from a library of `good' candidate paths $\PathSet = \seq{\Path}{\numRegion}$. 

Let $\hypSet = \set{\hyp_1, \dots, \hyp_\numHyp}$ be a set of ``hypotheses'', each of which is analogous to a world. We have a prior distribution $P(\hyp)$ on this set. A ``test'' $\test \in \testSet$ is performed by querying a corresponding edge $\edge \in \edgeSet$ for evaluation, which returns a binary outcome $\outcome \in \{0, 1\}$ denoting if an edge is valid or not. Thus each hypothesis can be considered a function, $\hyp: \testSet \rightarrow \{0, 1\}$, mapping tests to corresponding outcomes. The cost of performing a test is $\cost(\test)$. 
A path $\Path_i \in \PathSet$ corresponds to a set of worlds on which that path is valid. 
Hence each path $\Path_i \in \PathSet$ corresponds to a ``decision region'' $\region_i \subseteq \hypSet$ over the space of hypotheses.
Let $\set{\region_i}_{i=1}^{\numRegion}$ be the set of ``decision regions'' corresponding to $\PathSet$.

For a set of tests $\selTestSet \subseteq \testSet$ that are performed, let the observed outcome vector be denoted by $\obsOutcome$. Let the version space $\versionSpace(\obsOutcome)$ be the set of hypotheses consistent with outcome vector $\obsOutcome$, i.e. $\versionSpace(\obsOutcome) = \setst{\hyp \in \hypSet}{ \forall \test \in \selTestSet, \hyp(\test) = \obsOutcome(\test)}$.

We define a policy $\policy$ as a mapping from the current outcome vector $\obsOutcome$ to the next test to select. A policy terminates when at least one region is valid, or all regions are invalid. Let $\hyp$ be the underlying world on which it is evaluated. Denote the outcome vector of a policy $\policy$ as $\obsOutcomeFunc{\policy}{\hyp}$. The expected cost of a policy $\policy$ is $\cost(\policy) = \expect{\hyp}{\cost(\obsOutcomeFunc{\policy}{\hyp}}$ where $\cost(\obsOutcome)$ is the cost of all tests $\test \in \selTestSet$. The objective is to compute a policy $\policyOpt$ with minimum cost that ensures at least one region is valid, i.e.
\begin{equation}
\label{eq:drd}
\policyOpt \in \argminprob{\policy} \;\cost(\policy) \; \mathrm{s.t} \; \forall \hyp , \exists \region_d \; : \; P(\region_d \;|\; \obsOutcomeFunc{\policy}{\hyp}) = 1
\end{equation}

An illustration of this equivalence is shown in Fig.~\ref{fig:general_drd_problem}.

%% file: related_work.tex
\section{Related Work} \vspace{-0.7em}
\label{sec:related_work}
The computational bottleneck in motion planning varies with problem domain and that has led to a plethora of planning techniques (\cite{Lav06}). When vertex expansions are a bottleneck, A*~\cite{hart1968formal} is optimally efficient while techniques such as partial expansions~\cite{yoshizumi2000partial} address graph searches with large branching factors. However, we examine the problem class that is of particular importance in robotics - expensive edge evaluation. This is primarily because evaluation is performed by querying an underlying representation of the world that is built online and requires expensive geometric intersection computation. 

The problem class we examine, that of expensive edge evaluation, has inspired a variety of `lazy' approaches. The LazyPRM algorithm~\cite{bohlin2000path} only evaluates edges on the shortest path while FuzzyPRM~\cite{nielsen2000two} evaluates paths that minimize probability of collision. The Lazy Weighted A* (LWA*) algorithm~\cite{cohen2015planning} delays edge evaluation in A* search and is reflected in similar techniques for randomized search~\cite{gammell2015batch,Choudhury_2016_8070,hauser2015lazy}. An approach most similar in style to ours is the LazyShortestPath (LazySP) framework~\cite{dellin2016unifying} which examines the problem of which edges to evaluate on the shortest path. Instead of the finding the shortest path, our framework aims to efficiently identify a feasible path in a library of `good' paths. The Anytime Edge Evaluation (AEE*) framework~\cite{narayanan2017heuristic} also deals with a similar problem however it makes an independent edge assumption. Finally, there is a lot of work on modelling belief over configuration spaces \cite{choudhury2016pareto, pan2013faster, huh2016learning, burns2005sampling}. Using such models in DRD would be interesting future work.

We draw a novel connection between motion planning and optimal test selection which has a wide-spread application in medical diagnosis~\cite{kononenko2001machine} and experiment design~\cite{chaloner1995bayesian}. Optimizing the ideal metric, decision theoretic value of information~\cite{howard1966information}, is known to be NP\textsuperscript{PP} complete~\cite{krause2009optimal}. For hypothesis identification (known as the Optimal Decision Tree (ODT) problem), Generalized Binary Search (GBS)~\cite{dasgupta2004analysis} provides a near-optimal policy. For disjoint region identification (known as the Equivalence Class Determination (ECD) problem), \ecsq~\cite{golovin2010near} provides a near-optimal policy. When regions overlap (known as the Decision Region Determination (DRD) problem), HEC~\cite{javdani2014near} provides a near-optimal policy. The \direct algorithm~\cite{chen2015submodular}, a computationally more efficient alternative to HEC, forms the basis of our approach. We also employ the \bisect algorithm~\cite{choudhury2017active}, which solves the DRD problem under edge independence assumptions.

%% file: approach.tex
\section{Approach} \label{sec:approach}

\subsection{Overview}
\begin{figure}[!t]
    \centering
    \includegraphics[page=1,width=\columnwidth]{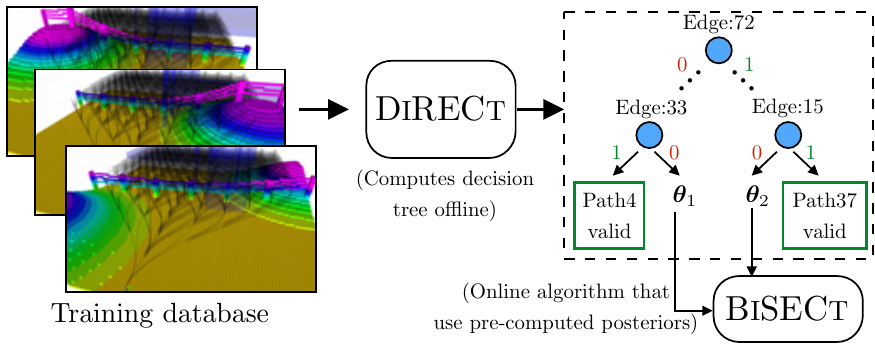}
    \caption{%
    \label{fig:framework}
    The overall approach framework. A training database is created by randomly sampling worlds from a generative model, collision checking the edge of the graph on each such world and creating a library of paths. The algorithm \direct is invoked to compute a decision tree offline. Each node of the tree contains the index of the edge to evaluate and branches on the outcome. The leaf node $i$ of the tree correspond either to a feasible path existing or the number of consistent worlds dropping below a threshold fraction $\threshDIRECT$. In the latter case, the bias vector $\biasVec_i$ is stored. 
	At test time, the tree is executed till a leaf node $i$ is reached. If the problem is unsolved at that point, the \bisect algorithm is invoked with $\biasVec_i$ as bias term. 
    }
\end{figure}

Fig.~\ref{fig:framework} shows an overview of our approach.
We sample a finite database of worlds to create a training dataset. 
We employ a greedy yet near-optimal algorithm \direct~\cite{chen2015submodular} to solve the DRD problem.
\direct chooses decisions to prune of inconsistent worlds from the database till it can ascertain if a path is valid. 
The decisions of \direct can be compactly stored in the form of a decision tree which is computed offline. 
At test time, the tree is executed till the leaf node is reached.
At this point, either the problem is solved or the fraction of consistent worlds drops below a threshold $\threshDIRECT$, i.e. it is likely that the test world is not in the database. 
In the latter case, we invoke another DRD algorithm, \bisect. 
\bisect implicitly reasons about the exhaustive set of $O(2^\edgeSet)$ worlds and does this efficiently by assuming edges are independent. 
\bisect is invoked with a bias vector of edge likelihoods $\biasVec$ computed from the remaining consistent worlds in \direct. 
The combined behaviour of the framework is as follows - the tree makes a set of evaluations to quickly collapse the posterior on to a set of candidate paths, while \bisect completes the episode being guided by the obtained posterior. We describe each component of the framework in the remaining subsections. 

\subsection{The Decision Region Edge Cutting algorithm (\direct) }
In order to solve the DRD problem in (\ref{eq:drd}), we adopt the framework of \emph{Decision Region Edge Cutting (\direct)}~\cite{chen2015submodular}. The intuition behind the method is as follows - as tests are performed, hypotheses inconsistent with test outcomes are pruned away. Hence, tests should be incentivized to push the probability mass over hypotheses into a region as fast as possible. Chen \etal~\cite{chen2015submodular} derive a surrogate objective function that not only provides such an incentive, but also exhibits  the property of adaptive submodularity~\cite{golovin2011adaptive} - greedily maximizing such an objective results in a near-optimal policy.

\direct uses a key result from Golovin \etal \cite{golovin2010near} who address the \emph{Equivalence Class Determination} (ECD) problem - a special case of the DRD problem (\ref{eq:drd}) when \emph{regions are disjoint}. Let $\{ \region_1, \dots, \region_\numRegion \}$ be a set of disjoint regions, i.e, $\region_i \cap \region_j = 0$ for $i \neq j$. Golovin \etal \cite{golovin2010near} provide an efficient yet near-optimal approach for solving ECD in their \ecsq algorithm. 
The \ecsq algorithm defines a graph $\graphEC=(\vertexEC, \edgeSetEC)$ where the nodes are hypotheses and edges are between hypotheses in different decision regions $\edgeSetEC = \cup_{i \neq j} \setst{ \edgeFnEC{\hyp}{\hyp'} }{\hyp \in \region_i, \hyp' \in \region_j}$. The weight of an edge is defined as $\weight(\edgeFnEC{\hyp}{\hyp'}) = P(\hyp) P(\hyp')$. 
An edge is said to be `cut' by an observation if either hypothesis is inconsistent with the observation. Hence a test $\test$ with outcome $\outcomeTest{\test}$ is said to cut a set of edges $\edgeSetEC(\outcomeTest{\test}) = \setst{\edgeFnEC{\hyp}{\hyp'}}{\hyp(\test) \neq \outcomeTest{\test} \vee \hyp'(\test) \neq \outcomeTest{\test}}$. The aim is to cut all edges by performing test while minimizing cost. 

\ecsq employs a weight function over regions, $\wec(\{\region_i\}) = \sum\limits_{i \neq j} P(\region_i) P(\region_j)$. Naively, computing the total edge weight requires enumerating all pairs of regions. However, we can compute this efficiently in linear complexity as $\wec(\{\region_i\}) = \frac{1}{2} \left( ( \sum\limits_i P(\region_i) )^2 - \sum\limits_i P(\region_i)^2 \right)$. 
\ecsq defines an objective function $\fec{\obsOutcome}$ that measures the ratio of the original weight of subregions $\region_i$ and the weight of pruned subregions $\region_i \cap \hypSpace(\obsOutcome)$, i.e. 
\begin{equation}
\label{eq:ec2_fn}
\fec{\obsOutcome} = 1 - \frac{\wec(\{\region_i\} \cap \hypSpace(\obsOutcome))}{\wec(\{\region_i\})}
\end{equation}

\begin{algorithm}[t]
  \caption{ \direct $\left( \activeHypSet, \hypRegion, \hypTest, \costTest \right)$ }\label{alg:direct}
  \algorithmStyle
  \For{$ \test \in \testSet$}
  {
    $\gainTest(\test) \gets 0$\;
    \For{$ \outcomeTest{\test} \in \{0, 1\} $} 
    {
      $\condActiveHypSet \gets \setst{\hyp \in \activeHypSet}{\hypTest(\hyp, \test)=\outcomeTest{\test}}$ \Comment*[r]{Prune hyp}
      $\probTest \gets \frac{\abs{\condActiveHypSet}}{\abs{\activeHypSet}}$ \Comment*[r]{Probability of outcome}
      $\gainTest(\test) \gets \gainTest(\test) + \probTest \; \mathtt{GainDRD} (\condActiveHypSet, \hypRegion)$\;
    }
    $\gainTest(\test) \gets \frac{\gainTest(\test)}{\costTest(\test)}$\;
  }
  \Return{$\argmaxprob{\test \in \testSet} \; \gainTest(\test)$}\;
\end{algorithm}

\ecsq uses the fact that $\fec{\obsOutcome}$ is \emph{adaptive submodular} (\cite{golovin2011adaptive}) to define a greedy algorithm. Let the expected marginal gain of a test be 
$\gain{\ec}{\test}{\outcome} = \expect{\outcomeTest{\test}}{ \fec{ \obsOutcomeAdd{\test} } - \fec{\obsOutcome} \;|\; \obsOutcome}$. \ecsq greedily selects a test $\test^* \in \argmaxprob{\test} \frac{\gain{\ec}{\test}{\obsOutcome}}{\cost(\test)}$.

We now return to the general DRD problem where regions are not disjoint. \direct reduces the DRD problem with $\numRegion$ regions to $\numRegion$ instances of the ECD problem. Each ECD problem is a `one region versus all'. ECD problem $i$ is defined over the following disjoint regions: the first region is $\region_i$ and the remaining regions are singletons containing only one hypothesis $\hyp \notin \region_i$. The \ecsq objective corresponding to this problem is $\feci{r}{\obsOutcome}$. The key idea is that \emph{solving any one ECD problem solves the DRD problem}. The \direct algorithm then combines them in a \emph{Noisy-OR} formulation by defining the following combined objective
\begin{equation}
\label{eq:drd_fn}
  \fdrd{\obsOutcome} = 1 - \prod\limits_{r=1}^\numRegion (1 - \feci{r}{\obsOutcome})
\end{equation}
\direct uses the fact that $\fdrd{\obsOutcome}$ is also \emph{adaptive submodular} to greedily select a test $\test^* \in \argmaxprob{\test} \frac{\gain{\drd}{\test}{\obsOutcome}}{\cost(\test)}$.
For details on the theoretical guarantees and proofs, we refer the reader to \cite{chen2015submodular}.

\begin{algorithm}[t]
  \caption{ $\mathtt{GainDRD} \left( \hypSetTwo, \hypRegion \right)$ }\label{alg:gain_drd}
  \algorithmStyle
  $v \gets 1$\;
  \For{$i \in \{1, \dots, \numRegion\} $}
  {
    $v \gets v \; \left( \frac{\mathtt{WeightEC}(\hypSetTwo, \hypRegion, i)}{\mathtt{WeightEC}(\hypSet, \hypRegion, i)} \right)$ \Comment*[r]{Gain from each ECD}
  }
  \Return{$v$}\;
\end{algorithm}

\begin{algorithm}[t]
  \caption{  $\mathtt{WeightEC} \left( \hypSetTwo, \hypRegion, i \right)$ }\label{alg:weight_ec}
  \algorithmStyle
  $a \gets \sum\limits_{\hyp \in \hypSetTwo} \hypRegion(\hyp, i) $ \Comment*[r]{Number of hyp in region}
  $b \gets \abs{\hypSet} - a$ \Comment*[r]{Remaining hyp}
  \Return{$ \frac{1}{2 \abs{\hypSet}^2} \left( (a+b)^2 - a^2 - b\right)$} 
\end{algorithm}

To aid in implementation, we provide a pseudo-code for \direct. The pseudo-code is derived by expanding and simplifying $\gain{\drd}{\test}{\obsOutcome}$ which we omit for brevity. 
Alg.~\ref{alg:direct} describes the \direct policy. $\activeHypSet$ is the set of active hypotheses which have remained consistent so far with test outcomes. $\hypRegion \in \real^{\numHyp \times \numRegion}$ is a binary membership matrix where $\hypRegion(\hyp, r) = 1$ if $\hyp \in \region_r$. $\hypTest \in \real^{\numHyp \times \abs{\testSet}}$ is the test outcome matrix where $\hypTest(\hyp, \test) = \hyp(\test)$. $\costTest \in \real^{\abs{\testSet} \times 1}$ is a vector of test costs. Alg.~\ref{alg:direct} computes the expected gain for each test by computing $\condActiveHypSet$, the set of hypotheses conditioned on test outcomes, and picks the best test. Alg.~\ref{alg:gain_drd} computes the DRD gain for $\condActiveHypSet$ by taking a product of individual ECD gains. Alg.~\ref{alg:weight_ec} calculates the weight of the $i^{\mathrm{th}}$ ECD problem. The computational complexity of Alg.~\ref{alg:direct} is $\bigo{\abs{\testSet} \numRegion \numHyp }$. Speedups can be obtained by lazily evaluating gains and using graph coloring to reduce the number of ECD problems~\cite{chen2015submodular}. 
\input{table_icra}

\subsection{Creating an offline decision tree using \direct}
\direct needs access to the entire training database which can be prohibitive at runtime for storage and computational reasons. 
We circumvent this problem by computing a decision tree offline using \direct and storing it.
The nodes of the tree encode which edge to evaluate. The tree branches on the outcome of the evaluation. Note that the depth of the tree is bounded by $\log_2(\numHyp)$ as all leaf nodes must be consistent with the training database. The size is further bounded by the fact that the tree terminates on a leaf node when the uncertainty has been pushed onto one region.

\subsection{Executing \bisect from the leaf node}
As discussed in Section~\ref{sec:intro}, it is impractical to have a database large enough to encompass all possible worlds that can arise at test time. Hence, if we reach the leaf node of the tree and the problem is still unsolved, we need to execute an online algorithm that can run to completion by reasoning over the exhaustive set of worlds.  
We use the \emph{Bernoulli Subregion Edge Cutting (\bisect)} algorithm~\cite{choudhury2017active} as our online algorithm. 
\bisect addresses the DRD problem under the assumption that test outcomes are independent Bernoulli random variables.
It leverages this assumption to reduce computational complexity from $\bigo{2^E}$ to $\bigo{E}$ and hence can be easily executed online.

\bisect needs as input a bias vector which corresponds to the independent likelihood of an edge being free. 
Since \direct has made a set of decisions to collapse the posterior, albeit on a finite database, we wish to use this to inform \bisect. 
We do this by growing the \direct decision tree only till the version space $\versionSpaceDIRECT$ drops below a fraction $\threshDIRECT$ of consistent worlds, i.e. $\abs{\versionSpaceDIRECT} \leq \threshDIRECT \abs{\hypSet}$. 
This is then used to create a bias vector $\biasVec$ with a mixture term to ensure non-zero support for all plausible worlds. The bias term for a test $\test$ is 
\begin{equation}
  \biasVec(\test) = \mixDIRECT \; \frac{1}{\abs{\versionSpaceDIRECT}} \sum\limits_{\hyp \in \versionSpaceDIRECT} \hypTest(\hyp, \test) + (1 - \mixDIRECT) \; 0.5
\end{equation}
Using $\biasVec$ leads to a more informed \bisect as compared to directly invoking \bisect from the beginning using a bias vector computed from the training database. 

%% file: table_icra.tex
\begin{table*}[!htpb]
\small
\centering
\caption{Normalized cost (with respect to our approach) of different algorithms on different datasets (lower and upper bounds of $95\%$ C.I.) }
\begin{tabulary}{\textwidth}{LCCCCCCC}\toprule
      & {\bf \algLazySP} & {\bf \algLazySPSet} 	& {\bf \algMaxTally}   & {\bf \algSetCover}  & {\bf \algMVOI} & {\bf \bisect} & {\bf \direct +} \\ 
      & 				 &						&  					  & 					  & 				& 			 & {\bf \bisect \phantom{+}} \\ \midrule
\multicolumn{8}{c}{ {\bf 2D Geometric Planning: Variation across environments} }   \\
Forest    	& $(10.90, 18.48)$	& $(1.84, 3.02)$	& $(0.17, 0.40)$	& $(0.14, 0.51)$	& $(0.30, 0.55)$	& $(0.014, 0.20)$	& $\red(0.00, 0.00)$\\ 
OneWall   	& $(7.47, 16.01)$ 	& $(0.30, 0.71)$	& $(0.00, 0.30)$	& $(0.08, 0.34)$    & $(0.09, 0.36)$   	& $\red(-0.06, 0.22)$ & $\red(0.00, 0.00)$ \\ 
TwoWall   	& $(21.54, 26.68)$ 	& $(0.00, 0.21)$ 	& $(0.20, 0.92)$    & $(0.12, 0.58)$    & $(0.31, 0.56)$   	& $(0.00, 0.53)$    & $\red(0.00, 0.00)$ \\ 
MovingWall  & $(1.33, 3.01)$	& $(1.00, 1.54)$ 	& $(0.43, 1.17)$    & $(0.35, 0.91)$    & $\red(-0.03, 0.57)$   & $(0.11, 0.92)$  & $\red(0.00, 0.00)$ \\ 
Baffle   	& $(7.86, 11.26)$	& $(2.30, 3.83)$ 	& $(0.33, 1.06)$    & $(0.36, 0.74)$    & $(0.26, 0.89)$   & $(0.11, 0.55)$    & $\red(0.00, 0.00)$ \\ 
Maze   		& $(14.39, 19.66)$	& $(1.16, 1.81)$ 	& $(0.12, 0.34)$    & $(0.00, 0.17)$    & $(0.41, 0.87)$   & $(0.44, 0.76)$    & $\red(0.00, 0.00)$ \\ 
Bugtrap   	& $(7.40, 8.57)$	& $(2.74, 3.53)$	& $(0.51, 0.84)$    & $\red(-0.12, 0.54)$   & $\red(-0.12, 0.53)$  & $(0.43, 0.91)$    & $\red(0.00, 0.00)$ \\ 
\multicolumn{8}{c}{ {\bf 2D Geometric Planning (Baffle): Variation across path library size} }   \\
$\numRegion:200$  
			& $(8.67, 11.20)$ 	& $(1.73, 2.34)$ 	& $(0.32, 0.73)$   & $(0.38, 0.89)$    & $(0.47, 1.17)$   & $\red(-0.03, 0.58)$    & $\red(0.00, 0.00)$  \\ 
$\numRegion:977$
			& $(7.20, 10.10)$ 	& $(1.35, 2.92)$	& $(0.24. 0.38)$   & $(0.31, 0.63)$    & $(0.20, 0.79)$   & $(0.03, 0.34)$    & $\red(0.00, 0.00)$\\ 
\multicolumn{8}{c}{ {\bf SE(2) Nonholonomic Path Planning: Variation across environments} }   \\
OneWall   	& $(2.22, 4.18)$ 	& $(0.15, 0.57)$	& $(0.16, 0.48)$   & $\red(-0.11, 0.07)$   & $(0.00, 0.28)$   & $\red(-0.07, 0.12)$    & $\red(0.00, 0.00)$  \\ 
MovingWall  & $\red(-0.14, 0.23)$	& $\red(-0.14, 0.15)$ 	& $(0.24, 0.49)$  	& $(0.13, 0.41)$    & $(0.00 0.36)$   & $(0.10, 0.54)$    & $\red(0.00, 0.00)$\\ 
Baffle   	& $(7.74, 10.48)$	& $(2.88, 4.81)$ 	& $(1.86, 3.21)$    & $(1.35, 2.32)$    & $(0.70, 1.47)$   & $(1.14, 1.70)$    & $\red(0.00, 0.00)$ \\ 
Bugtrap   	& $(3.75, 6.51)$ 	& $(2.27, 4.69)$ 	& $(0.22, 0.52)$    & $(0.05, 0.43)$    & $(0.26, 0.55)$   & $(0.12, 0.44)$    & $\red(0.00, 0.00)$ \\ 
\multicolumn{8}{c}{ {\bf Autonomous Helicopter Path Planning: Variation across environments} }   \\
Wires   	& $(17.42, 75.85)$ 	& $(1.15, 3.08)$ 	& $(0.55, 0.96)$   	& $(0.00, 0.25)$    & $\red(-0.08, 0.08)$   & $(0.08, 0.23)$  & $\red(0.00, 0.00)$  \\ 
Canyon   	& $(0.73, 1.27)$ 	& $(1.41, 2.00)$ 	& $(0.15, 0.52)$    & $(0.07, 0.40)$    & $(0.43, 0.72)$   		& $(0.06, 0.47)$  & $\red(0.00, 0.00)$ \\
\multicolumn{8}{c}{ {\bf 7D Arm Planning: Variation across environments} }   \\
Clutter   	& $(0.49, 1.08)$	& $(0.09, 0.57)$	& $\red(-0.04, 0.05)$    & $(0.00, 0.13)$    & $(0.10, 0.32)$   & $(0.00, 0.10)$    & $\red(0.00, 0.00)$ \\ 
Table+Clutter  & $(0.94, 1.84)$	& $\red(-0.22, 0.17)$	& $(0.06,0.51)$    & $(0.05, 0.27)$    & $(0.11, 0.46)$   & $(0.06, 0.36)$    & $\red(0.00, 0.00)$ \\ 
\bottomrule
\end{tabulary}
\label{tab:benchmark_results}
\end{table*}

%% file: experiments.tex
\section{Experiments} 
\label{sec:experiments}

\subsection{Dataset construction}
\label{sec:experiments:dataset}
We evaluated our approach on a collection of datasets spanning a spectrum of motion planning applications that range from simplistic yet insightful 2D problems to more realistic high dimension problems as encountered by a helicopter or a robot arm. The autonomous helicopter dataset in particular is our target application.
A typical dataset is constructed as follows. The robot dynamics information is used to create an explicit graph $\explicitGraph = \pair{\vertexSet}{\edgeSet}$ and a start and goal vertex. A dataset of $\numHyp$ worlds is sampled from a designed generative model. Each edge is evaluated on each world to create a test outcome matrix $\hypTest \in \real^{\numHyp \times \abs{\testSet}}$. A library of paths is created by solving for $k-$shortest paths on the dataset and sub-sampling it to maintain a size of $\numRegion$. This is then used to create a binary membership matrix $\hypRegion \in \real^{\numHyp \times \numRegion}$ encoding the validity of a path on a world. $10\%$ of the data is used for test, remainder for training. The algorithms work with these abstract representations and do not need access to application specific details. Refer to Choudhury \etal~\cite{choudhury2017active} for more details on dataset construction. \footnote{Typical values used are $\numHyp:1000$, $\numRegion:500$. We plan to provide a link to open source code and datasets for the camera ready version.}

\begin{figure*}[t]
    \centering
    \includegraphics[page=1,width=\textwidth]{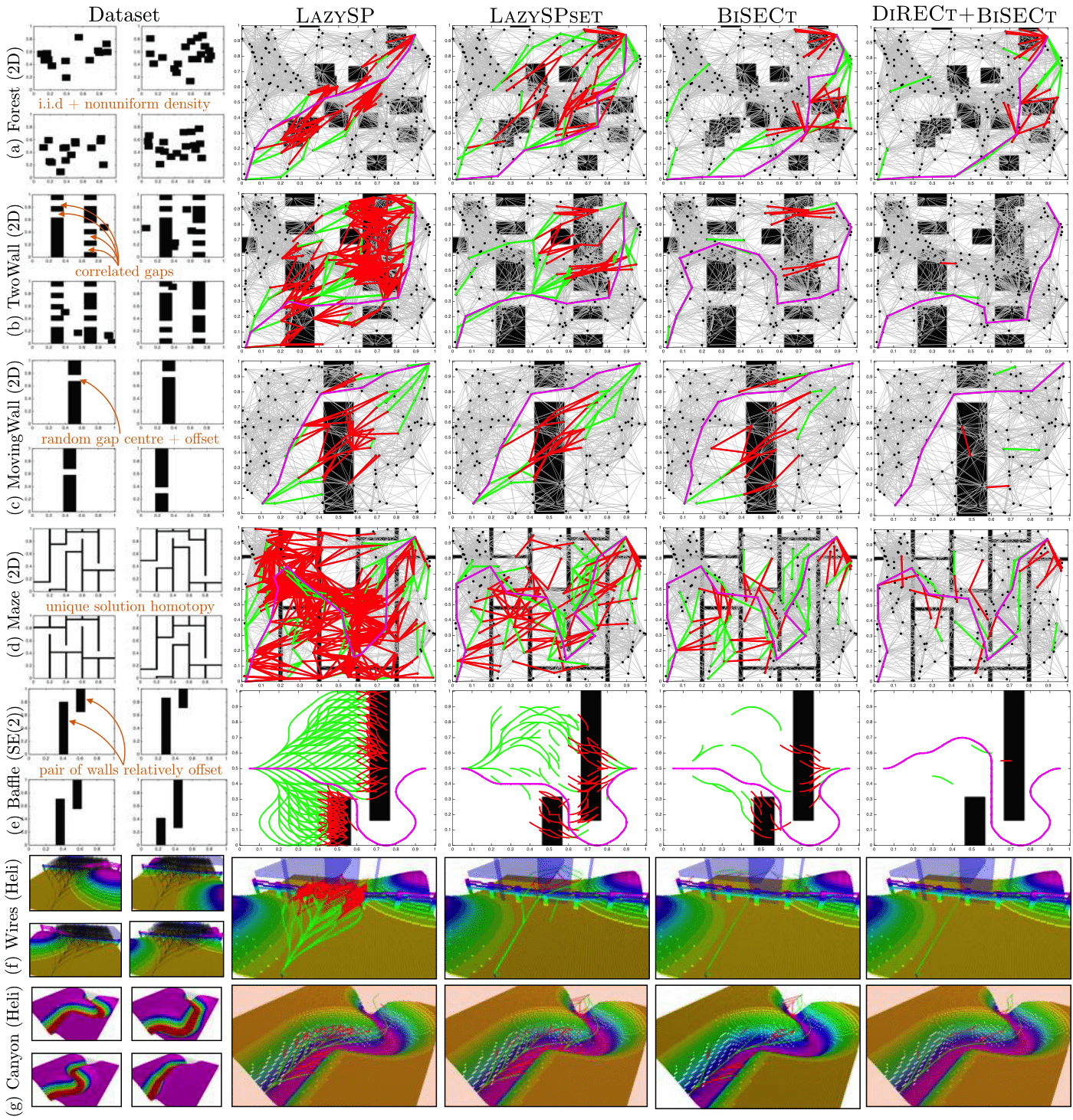}
    \caption{%
    \label{fig:algo_comparisons}
    Comparison of \algLazySP, \algLazySPSet, \bisect and \direct+\bisect on a selection of datasets. 4 samples from each dataset is shown. The final performance of all algorithms on a test problem is shown: valid edges checked (green) and invalid edges checked (red). \fullFigGap
    }
\end{figure*}%

\begin{figure*}[t]
    \centering
    \includegraphics[page=1,width=\textwidth]{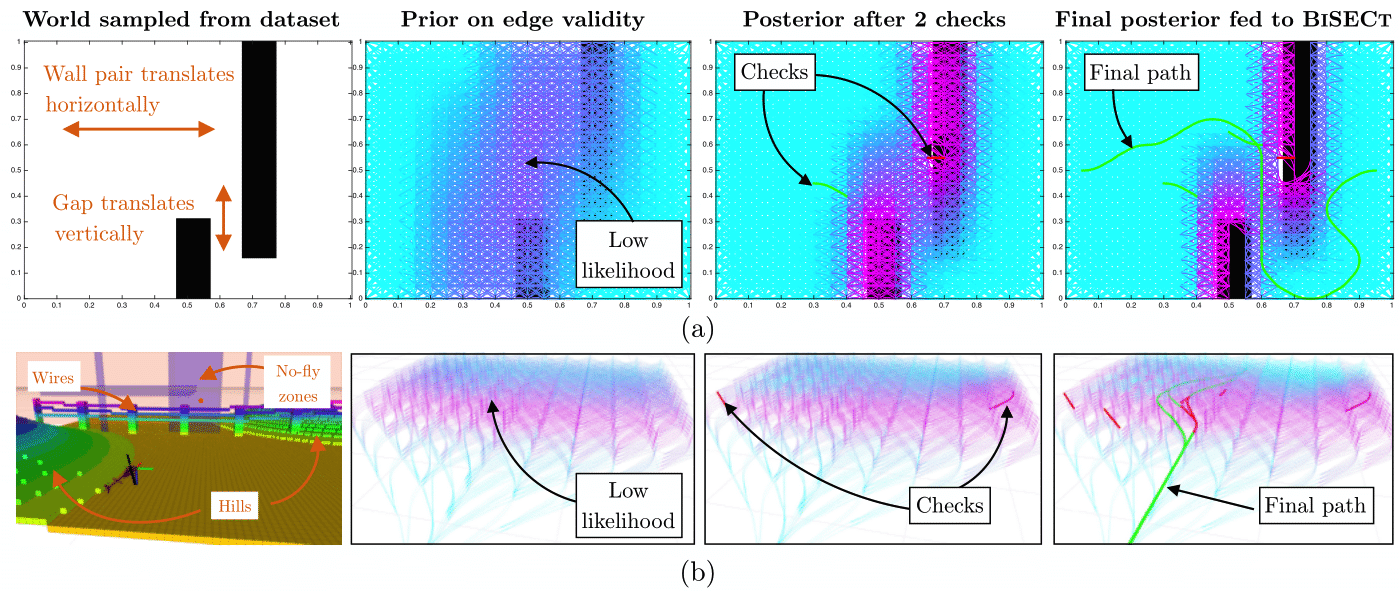}
    \caption{%
    \label{fig:belief_propagation}
    \direct performs edges evaluation to collapse the uncertainty about the validity of a path. (a) An example from the Baffle dataset for SE(2) nonholonomic path planning. Here two walls occur in a pair forcing the path to maneuver through the gap. The prior shows only a general location where obstacles are likely to occur. After 2 checks, \direct is able to locate the gap. The resultant posterior allows \bisect to finish off the episode.
    (b) A realistic example from the Wires dataset for autonomous helicopter path planning. Here the helicopter is flying over a terrain that may have powerlines. The terrain also has natural obstacles such as hills. Presence of other aircrafts and no-fly zones also require avoidance. The prior shows a band of low likelihood region that corresponds to the presence of the wires. After 2 checks, \direct is able to infer the location of obstacles on either flank. The resultant posterior allows \bisect to focus on the centre region and find a path. \fullFigGap
    }
\end{figure*}%

\subsection{Baseline algorithms}
\label{sec:experiments:baseline}
Our primary baseline is \bisect~\cite{choudhury2017active} which treats each edge as independent Bernoulli random variables (i.e. averages $\hypTest$ along each column to use as bias). We additionally use high performing baselines from Choudhury \etal~\cite{choudhury2017active} which were competitive with \bisect, i.e the \algMaxProbReg version of \algMaxTally, \algSetCover and \algMVOI. We add to this the \algLazySP algorithm~\cite{dellin2016unifying} which operates on the original graph $\explicitGraph$. We also introduce a new algorithm \algLazySPSet which is restricted to the library of paths $\PathSet$.

\subsection{Summary of results}
\label{sec:experiments:discussion}
Table~\ref{tab:benchmark_results} shows the evaluation cost of all algorithms on various datasets normalized w.r.t \direct+\bisect. The two numbers are lower and upper $95\%$ confidence intervals - hence it conveys how much fractionally poorer algorithms are w.r.t our approach. The best performance on each dataset is highlighted. 
Fig.~\ref{fig:algo_comparisons} shows a comparison of algorithms on certain datasets. 
We present a set of observations to interpret these results.
\begin{observation}
\direct+\bisect has a consistently competitive performance across all datasets.
\end{observation} \vspace{-0.5em}

Table~\ref{tab:benchmark_results} shows on $17$ datasets, \direct is at par with the best - on $8$ of those it is exclusively the best. 

\begin{observation}
\direct is more effective on environments with spatial correlation. 
\end{observation} \vspace{-0.5em}

Fig.~\ref{fig:algo_comparisons} shows that datasets such as TwoWall, MovingWall, Maze and Baffle are more structured. For example in the Maze dataset, there are 5 hallways with one interconnecting passage. \direct is able to locate this passage with a few checks and has better performance than \bisect which assumes independence between edges.
On the other hand the Forest dataset has less spatial correlation and \bisect performs comparably (has an upper margin of $0.20$).
Similar phenomemnon was observed in 7D arm planning between Clutter (less correlation) and Table+Clutter (more correlation) datasets.

\begin{observation}
\direct+\bisect improves in performance with more data.
\end{observation} \vspace{-0.5em}

\begin{figure}[t]
    \centering
    \includegraphics[width=\columnwidth]{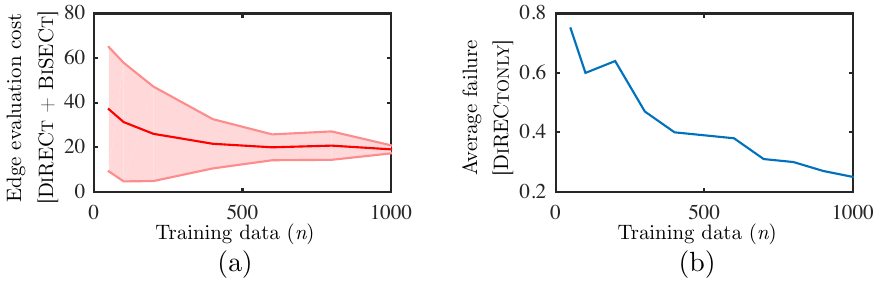}
    \caption{%
    \label{fig:analysis_plot}
    (a) Mean and variance of edge evaluation cost of \direct+\bisect with increasing training size. (b) The average failure (to indentify a feasible path) rate when only using \direct (without \bisect). \fullFigGap
    }
\end{figure}

Fig.~\ref{fig:analysis_plot}(a) shows that both mean and variance reduce as the size of the dataset is increased. This is not only due to \direct having better realizability, but also due to \bisect having a more accurate bias term.

\begin{observation}
\bisect is essential as a post-processing step
\end{observation} \vspace{-0.5em}

We defined an algorithm, \directONLY that runs \direct to completion and randomly returns a path from the consistent set of paths, i.e. the a path \direct believes should be feasible. Fig.~\ref{fig:analysis_plot}(b) shows the failure rate of \directONLY with training size, i.e. the returned path being infeasible. The plot shows the failure does not go to zero. \bisect is essential to reason about the remaining paths and in which order to check edges to ascertain which path is free.

\subsection{Case study: Roles played by \direct and \bisect}
We take a closer look at the Baffle dataset for SE(2) path planning as shown in Fig.~\ref{fig:belief_propagation}(a). The combination of the narrow gap between two walls and the curvature constraint of the robot makes this a challenging problem as shown by the performance of baseline LazySP in Fig.~\ref{fig:algo_comparisons}(e). Also note that \bisect too struggles on this problem. Returning back to Fig.~\ref{fig:belief_propagation}(a), we see that the prior over edge validity is not informative enough for \bisect to find the gap. As \direct proceeds to collision check edges, it is quickly able to localize the gap between the two walls. Interestingly, it is \emph{relatively uncertain about the actual vertical location of the wall} - this is reflective of \direct judiciously reducing uncertainty only enough to make a region valid (i.e to know if a candidate path would be feasible). The posterior is much more informative for \bisect which is able to easily find a feasible path.

We see a similar phenomenon in the Wires dataset for helicopter planning in Fig.~\ref{fig:belief_propagation}(b). As \direct proceeds to collision check edges, it is quickly able to ascertain presence of hills in the two flanks and a gap in the centre. \bisect uses this posterior to focus along the centre and find a path.

%% file: conclusion.tex
\section{Discussion and Future Work}
\label{sec:conclusion}
In this paper, we addressed the problem of identification of a feasible path from a library while minimizing the expected cost of edge evaluation given priors on the likelihood of edge validity. We showed that this problem is equivalent to a DRD problem where the goal is to select tests (edges) that drive uncertainty into a single decision region (a valid path). 
We proposed an approach that combines two DRD algorithms, \direct and \bisect, to efficiently solve the problem.
We validated our approach on a spectrum of problems against state of the art heuristics and showed that it has a consistent performance across datasets. 
These results demonstrate the efficacy of leveraging prior data to significantly reduce collision checking effort. 
We now discuss some insights and directions for future work.

\begin{ques}
How can we relax the restrictions in the framework in (\ref{sec:problem_formulation}) - the prior is specified only via a finite database of worlds and selection is limited to a fixed library of paths.
\end{ques} \vspace{-0.5em}

An alternate approach to modeling belief over configuration spaces is to assume edges are locally correlated. Under this assumption, one can use local models such as KDE~\cite{choudhury2016pareto}, mixture of Gaussians~\cite{huh2016learning}, RKHS~\cite{ramos2016hilbert} or even customized models~\cite{pan2013faster}. 
The efficacy of these models depends on how accurately they can represent the world, how efficiently they can be updated and how efficiently they can be projected on the graph.
The active learning not only needs to reason about the current belief of the world, but belief posteriors conditioned on possible outcomes of edge evaluation. 

Explicitly reasoning about a set of paths is expensive as the size of the set can be exponential in the number of edges in the graph. An alternate method is to directly reason about a distribution over all possible paths between two vertices implicitly, however, this can be intractable. Tractable approximations to such functions have been explored in the context of edge selection~\cite{dellin2016unifying}. Adopting such techniques in the active learning setting would be interesting to pursue.

\begin{ques}
We have so far been concerned with finding a feasible path. Can we extend our framework to the optimal path identification problem?
\end{ques} \vspace{-0.5em}

Introducing an additional criteria of minimizing path cost creates a tension between producing high quality paths and expending more evaluation effort. A desirable behaviour is to have an anytime algorithm that traverses the Pareto-frontier~\cite{choudhury2016pareto,choudhury2017densification}. We can tweak our algorithm to display such behavior - we first solve the feasible path identification problem, prune all costlier paths (including this) from the library, prune worlds which belonged only to those paths, and then solve the feasible path problem again. However, while this will eventually converge to the optimal path, we can not necessarily control the speed of convergence.